\documentclass{article}
\usepackage{spconf,amsmath,amssymb,graphicx}

\title{GAIT: GRADIENT ADJUSTED UNSUPERVISED IMAGE-TO-IMAGE TRANSLATION}

\name{Ibrahim Batuhan AKKAYA\textsuperscript{1,2}, Ugur HALICI\textsuperscript{2,3}}
\address{   \textsuperscript{1}Research Center, Aselsan Inc., Turkey\\
            \textsuperscript{2}Department of Electrical and Electronics Engineering, Middle East Technical University, Turkey\\
            \textsuperscript{3}NOROM Neuroscience and Neurotechnology Excellency Center, Turkey}
\begin{document}
%
\maketitle
\begin{abstract}
Image-to-image translation (IIT) has made much progress recently with the development of adversarial learning. In most of the recent work, an adversarial loss is utilized to match the distributions of the translated and target image sets. However, this may create artifacts if two domains have different marginal distributions, for example, in uniform areas. In this work, we propose an unsupervised IIT method that preserves the uniform regions after the translation. The gradient adjustment loss, which is the L2 norm between the Sobel response of the target image and the adjusted Sobel response of the source images, is utilized. The proposed method is validated on the jellyfish-to-Haeckel dataset, which is prepared to demonstrate the mentioned problem, which contains images with different background distributions. We demonstrate that our method obtained a performance gain compared to the baseline method qualitatively and quantitatively, showing the effectiveness of the proposed method.
\end{abstract}

{\let\thefootnote\relax\footnote{{©2020  IEEE. Personal use  of  this  material  is  permitted.   Permission from IEEE must be obtained for all other uses, in any current or future media,including reprinting/republishing this material for advertising or promotional purposes, creating new collective works, for resale or redistribution to servers or lists, or reuse of any copyrighted component of this work in other works. This paper has been accepted by the 27th IEEE International Conference on Image Processing (ICIP 2020).}}}

\section{Introduction}
\label{sec:intro}

The purpose of the image-to-image translation (IIT) problem is to learn a mapping from the source domain to the target domain. Many computer vision problems such as semantic segmentation, super-resolution, image coloring, and single image depth estimation can be considered as IIT problems. For example, the semantic segmentation problem aims to group the pixels of semantic elements such as cars and roads in the image. The class of the object corresponding to each pixel in the image is tried to be estimated. This approach can also be considered as a transformation from the image to the class label map.


Methods such as DualGAN\cite {yi2017dualgan}, DiscoGAN\cite {kim2017learning}, and CycleGAN \cite{zhu2017unpaired} use adversarial training for both transformation from source to target domain and from target to source domain. Besides, the preservation of the original image is encouraged by the cyclic consistency loss function when transforming from source to target and then back to the source. The same loss function is valid when converting from target to source and then back to the target. Cyclic consistency has played an essential role in increasing translation success.

In this study, the CycleGAN method is taken as the baseline method. The CycleGAN method may create artifacts during the translation of the uniform background of the image in the source domain to the target domain. In this study, the gradient of the source image is preserved after the translation in order to solve the mentioned problem. The gradient estimation is obtained with the Sobel operators, and the gradient adjustment loss function is defined on Sobel responses. In addition to that, an adjustment constant is defined to scale gradient between source and target domain, which increases the quantitative performance. The contributions of this study can be listed as follows.

\begin{itemize}
    \item By defining the gradient adjustment loss function, the gradient is forced to be preserved after the translation. This loss function contributes to preventing background distortion.
    \item The gradient of the source image is boosted or reduced by gradient adjustment constant that increases the quantitative performance.
    \item Jellyfish-to-Haeckel dataset is prepared. The efficiency of the proposed method is shown on this dataset.
\end{itemize}

In the second part of the paper, the studies on the proposed method are mentioned. The proposed method, together with the loss functions, is explained in the third section. In the fourth section, details about the implementation are given. In the fifth section, datasets are explained, qualitative and quantitative results are given. The paper is concluded in the last section.

\section{Related works}

\subsection{Generative Adversarial Networks}
Generative adversarial networks (GAN), first proposed by Goodfellow et al. \cite{goodfellow2014generative}, are frequently used in the solution of the IIT problem. The success of the GANs in generating realistic images has made these networks attractive to the IIT problem. A fully connected neural network is used in \cite{goodfellow2014generative}, and it is evaluated on relatively simple image datasets. However, it has shown limited performance in high-resolution image generation. Radford et al. \cite{radford2015unsupervised} proposed the deep convolutional neural network (DCGAN), adapting the network architecture to the convolutional neural networks in the pioneering study. Deconvolution layers in DCGAN enabled generative networks to generate images at higher resolutions. Mao et al. \cite{mao2017least} stated that the cross-entropy loss functions used in the mentioned methods might cause gradient loss problem. They aimed to solve the problem by using the least-squares loss function instead of the cross-entropy.

\subsection{Image-to-image Translation}
IIT problem can be studied with supervised and unsupervised learning approaches. Pix2Pix method\cite{isola2017image} is an example of a supervised IIT method. The method uses conditional GAN, which is conditioned on to the reference image. L1 loss between the reference and the generated image is utilized.

The CycleGAN \cite{zhu2017unpaired} method took a similar approach. This method defines the cyclic consistency loss function in addition to GAN loss, eliminating the need for a paired dataset. DualGAN\cite{yi2017dualgan} and DiscoGAN\cite{kim2017learning} methods, which are proposed concurrently with CycleGAN, use the same principle, although they use different loss functions. 

These methods transform the entire image. For this reason, artifacts may occur in some parts of the generated image. For the solution of this problem, Mejjati et al. \cite{mejjati2018unsupervised} proposed a method that determines the parts that need to be transformed on the image by using the attention mechanism in an unsupervised manner. However, this method does not make any modification to the parts where the attention map is zero, which may cause that the translated image might not match the target domain.

Some IIT methods recognize that there is a common latent space between the source and the target domain. CoGAN\cite{liu2016coupled} aims to use this common space by sharing parameters between the generator and discriminator networks of the source and target domain. Similarly, the UNIT\cite{liu2017unsupervised} method converts between domains by learning features in low-dimensional shared latent space. The MUNIT\cite{huang2018multimodal} method extends the UNIT method and offers a solution to the multi-modal IIT problem.

\section{Proposed Method}

The purpose of unsupervised IIT is to estimate a mapping ($F_{S\xrightarrow{}T}$) without using paired images from the source ($S$) domain to the target ($T$) domain. This translation uses $X_S$ and $X_T$ image sets, which are sampled independently from the source and target domains, respectively. The image generated by translation from the source domain to the target domain should match the probabilistic distribution of the target domain. Mathematically, this expression can be expressed as follows: $F_{S\xrightarrow{}T}(X_S)\in{}P_T$. An adversarial loss function is utilized in order to satisfy matching.

The CycleGAN method, which we use as the baseline algorithm, learns the inverse translation $F_{T\xrightarrow{}S}$ in addition to the forward translation. It uses these two mappings to define a loss function to promote cyclic consistency that can be defined as follows: The image converted from the source domain to the target domain should be same as the original image when converted back to the source domain: $F_{T\xrightarrow{}S}(F_{S\xrightarrow{}T}(X_S))\simeq{}X_S$. This restriction should also be satisfied in the opposite translation.

\begin{figure*}[t]
\centering
    \scalebox{1.0}{\includegraphics{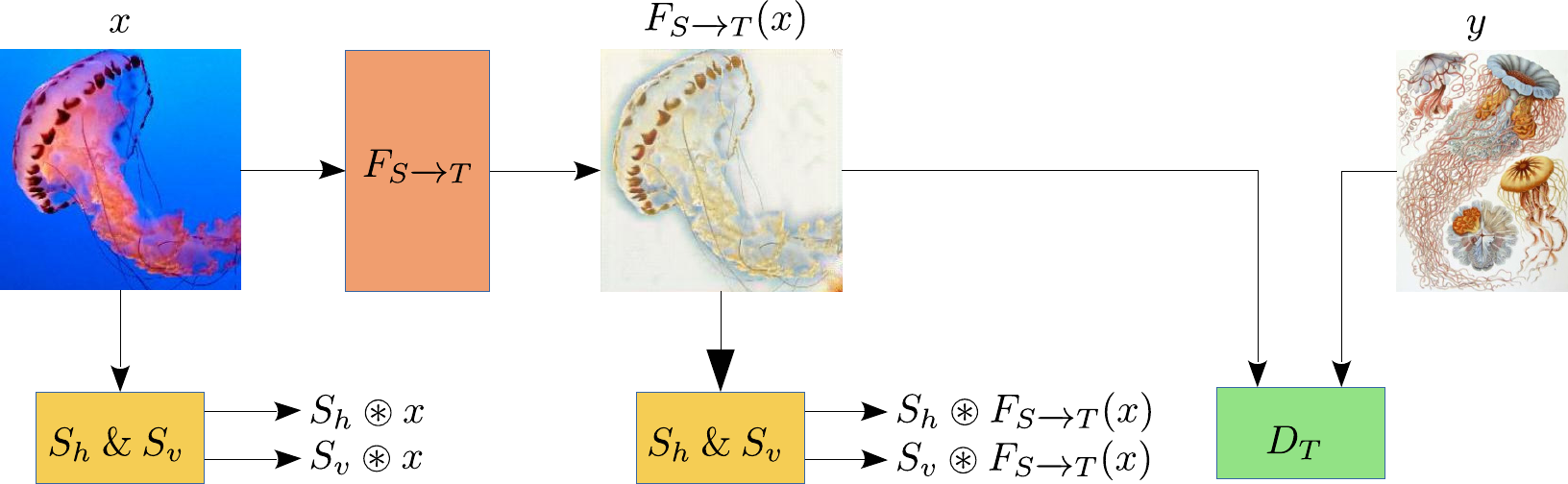}}
    \caption{The data-flow diagram of GAIT}
    \label{fig:sobelgan_overview}
\end{figure*}

In addition to the adversarial and cyclic consistency loss function, which are already used in the CycleGAN method, the gradient adjustment loss function proposed in this study is also described in the following subsections. The horizontal and vertical gradient estimation of the source and the target images are calculated by Sobel operators. The gradient adjustment loss function is defined in these estimations. The data-flow diagram of the proposed method from the source to the target domain is shown in Figure \ref{fig:sobelgan_overview}. An input image $x$ is translated by the generator network $F_{S\xrightarrow{}T}$. A copy of the diagram has also been used to convert in the opposite direction.

\subsection{Adversarial Loss Function}
When image translation is performed, the adversarial loss function is used to match the marginal distribution of the generated image to the marginal distribution of the target domain. In this study, the loss functions proposed in LSGAN\cite{mao2017least} method are used.

Adversarial loss function for generator network $\mathcal{L}_{advF}$ and adversarial loss function for discriminator network $\mathcal{L}_{advD}$ are given in \ref{eq:loss_adv_f} and \ref{eq:loss_adv_d} respectively. In these equations, $F_{S\xrightarrow{}T}$ refers to the generator network that translates from the source domain to the target domain, and $D_T$ refers to the discriminator network for the target domain. These loss functions also apply to networks that convert from the target to the source domain. The superscript in the equations represents the domain of the original image. 

\begin{equation}
    \label{eq:loss_adv_f}
    \mathcal{L}_{advF}^S = \mathbb{E}_{x\sim{}P_S}[(D_T(F_{S\xrightarrow{}T}(x))-1)^2]
\end{equation}

\begin{equation}
\label{eq:loss_adv_d}
\begin{split}
    \mathcal{L}_{advD}^S = \frac{1}{2} \mathbb{E}_{y\sim{}P_T}[(D_T(y)-1)^2]\\
    + \frac{1}{2} \mathbb{E}_{x\sim{}P_S}[D_T(F_{S\xrightarrow{}T}(x))^2]
\end{split}
\end{equation}

During the training, generator network parameters are updated to minimize $\mathcal{L}_{advF}$, and discriminator network parameters are updated to minimize $\mathcal{L}_{advD}$. 

\subsection{Cyclic Consistency Loss Function}
Adversarial learning promotes the generation of images that resembles the images of the target domain. To obtain more realistic images, in addition to the adversarial loss function, the cyclic consistency loss function has been defined. This function encourages the original image to be preserved when the image is translated back to the source domain after translated to the target domain. The L1 norm between the reconstructed image and the original image is defined as the loss function to achieve this goal. This statement is given mathematically in the equation \ref{eq:cyclic_consistency}. 

\begin{equation}
    \label{eq:cyclic_consistency}
    \begin{split}
    \mathcal{L}_{cyc} = \mathbb{E}_{x\sim{}P_S}[\left\lVert x-F_{T\xrightarrow{}S}(F_{S\xrightarrow{}T}(x))\right\rVert_1] \\+  \mathbb{E}_{y\sim{}P_T}[\left\lVert y-F_{S\xrightarrow{}T}(F_{T\xrightarrow{}S}(y))\right\rVert_1]
    \end{split}
\end{equation}

\subsection{Gradient Adjustment Loss Function}
Adversarial and cyclic consistency loss functions encourage the artificial neural network to create realistic visuals that resemble the images in the target domain. However, it can cause artifacts in images with a uniform background during translation. The gradient adjustment loss function has been proposed in this paper in order to prevent artifacts. 

The gradient adjustment loss uses the Sobel filter to estimate the image gradients. The Sobel filter consists of two kernels. When these two kernels, shown with $S_h$ and $Sy$, are convolved with images, they calculate derivative approximations in horizontal and vertical directions, respectively. $S_h$ and $Sy$ values are shown in equation \ref{eq:sobel_kernels}.

\begin{equation}
\label{eq:sobel_kernels}
\begin{split}
S_h = \begin{bmatrix}
-1 & 0 & 1 \\
-2 & 0 & 2 \\
-1 & 0 & 1
\end{bmatrix}, S_v = S_h^T
\end{split}
\end{equation}

The gradient adjustment loss can be defined as the L2 norm between the Sobel response of the target image and the adjusted Sobel response of the source images.
Sobel loss function is given in equation \ref{eq:sobel_loss} where $c_{ga}$ corresponds to the gradient adjustment constant.

\begin{equation}
    \label{eq:sobel_loss}
    \begin{split}
    \mathcal{L}_{grad} = \mathbb{E}_{x\sim{}P_S} 
    [\left\lVert (S_h \circledast{} x)c_{ga} - S_h \circledast{} F_{S\xrightarrow{}T}(x)\right\rVert_2^2 \\
    + \left\lVert (S_v \circledast{} x)c_{ga} - S_v \circledast{} F_{S\xrightarrow{}T}(x)\right\rVert_2^2] \\
    + \mathbb{E}_{y\sim{}P_T}
    [\left\lVert (S_h \circledast{} y)/c_{ga} - S_h \circledast{} F_{T\xrightarrow{}S}(y)\right\rVert_2^2 \\
    + \left\lVert (S_v \circledast{} y)/c_{ga} - S_v \circledast{} F_{T\xrightarrow{}S}(y)\right\rVert_2^2]   
    \end{split}
\end{equation}

\subsection{Total Loss Function}

The total loss functions for the generator and discriminator network are given in the equation \ref{eq:total_loss}. Both the generator and the discriminator network are trained to minimize the corresponding loss functions.

\begin{equation}
\label{eq:total_loss}
\begin{split}
    \mathcal{L}_{F} &= \mathcal{L}_{advF}^S + \mathcal{L}_{advF}^T + \lambda_{cyc} \mathcal{L}_{cyc} + \lambda_{grad} \mathcal{L}_{grad}\\
    \mathcal{L}_{D} &= \mathcal{L}_{advD}^S + \mathcal{L}_{advD}^T
\end{split}
\end{equation}

\section{Implementation}
\subsection{Neural Network Architectures}
The architectures in the CycleGAN method are used in the implementation of the proposed method. The neural network proposed by Johnson et al. \cite{johnson2016perceptual} is used as the generator network. 256 x 256 is preferred as the image resolution. PatchGAN network proposed by Isola et al. \cite{isola2017image} is used as the discriminator network.

\subsection{Training Details}
In all experiments, Adam optimizer is used. The $\lambda_{cyc}$, $\lambda_{grad}$ parameters and learning rate are selected as 10, 630 and 0.0001, respectively. All networks are initialized with random coefficients.

\section{Experiments}

\vspace{-10pt}
\subsection{Datasets}
\begin{figure}[]
\centering
    \scalebox{.25}{\includegraphics{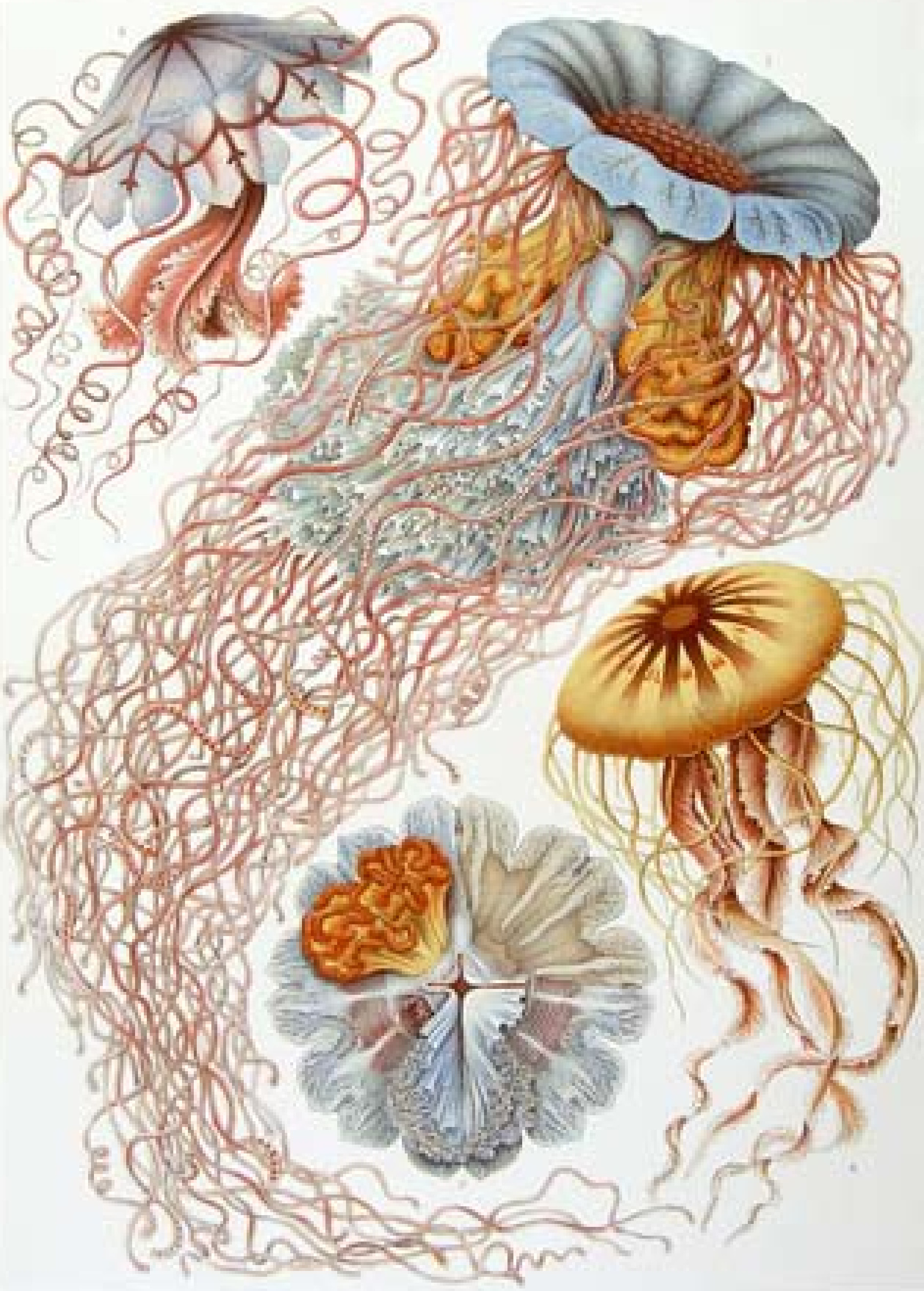}}
    \scalebox{.25}{\includegraphics{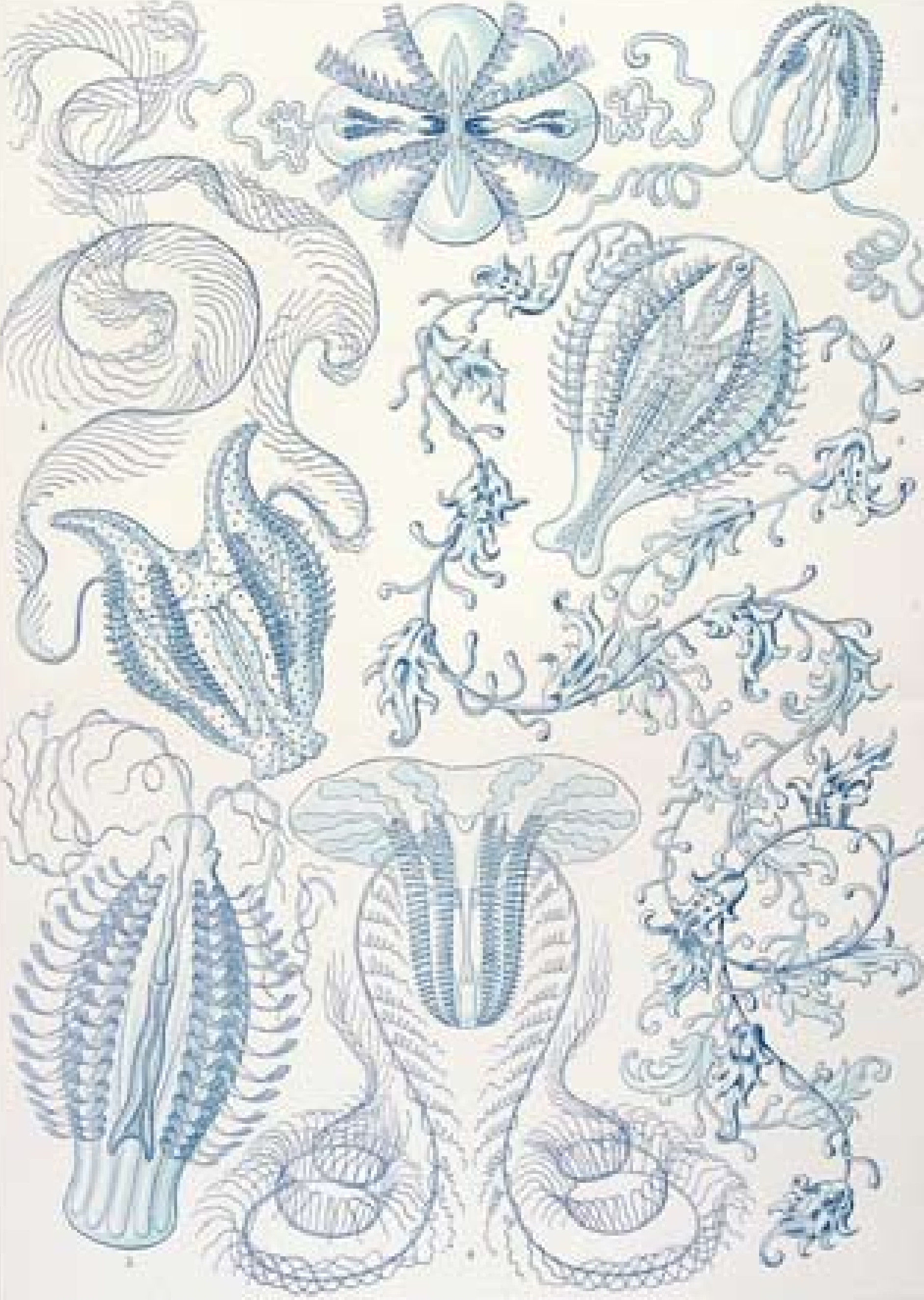}}
    \scalebox{.25}{\includegraphics{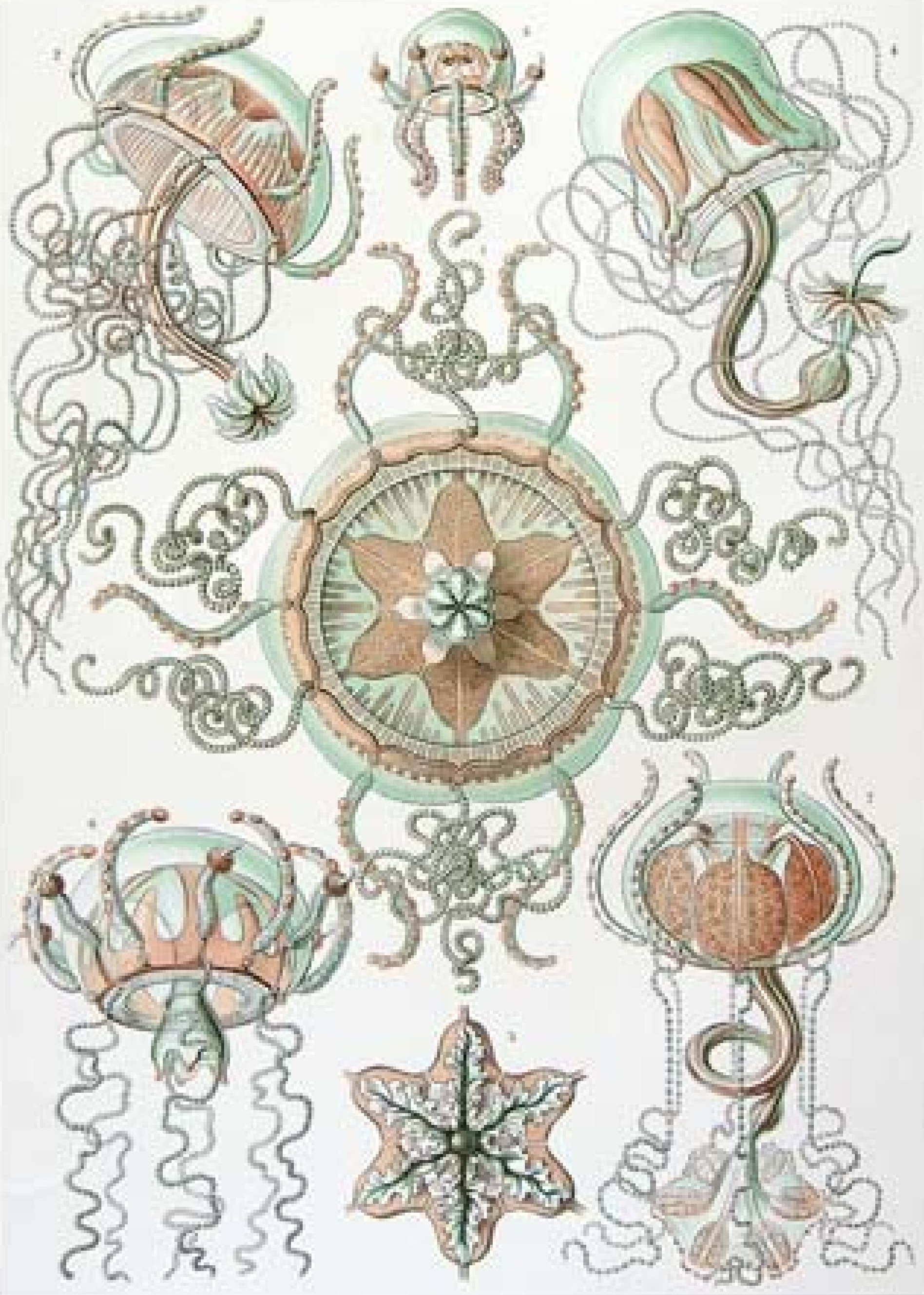}}
    \scalebox{.25}{\includegraphics{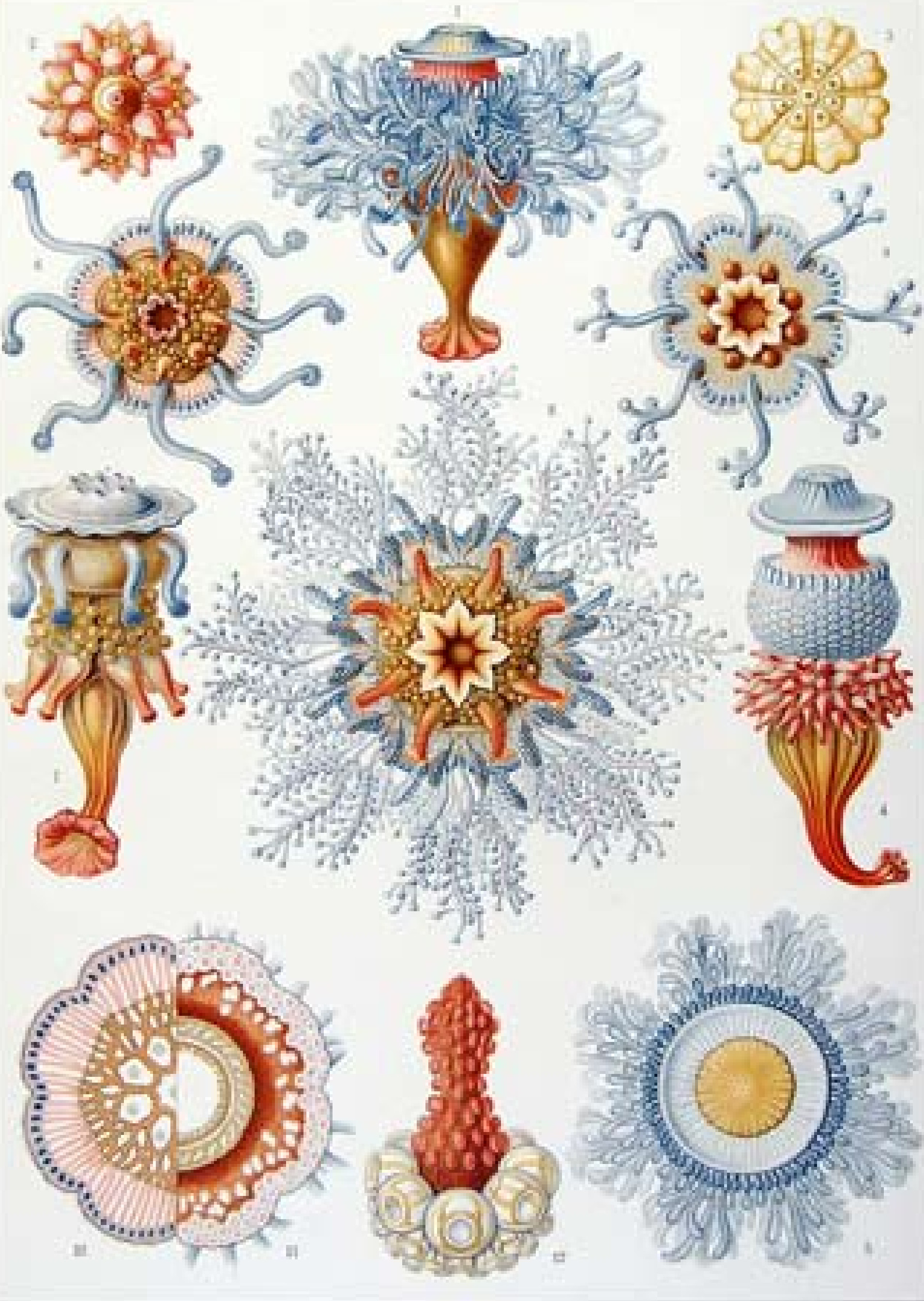}}
    \caption{Samples from Haeckel dataset}
    \label{fig:artform_samples}
\end{figure}

The proposed method is evaluated on two datasets, namely Aerial-to-Maps and Jellyfish-to-Haeckel. 
The jellyfish-to-Haeckel dataset consists of 2822 jellyfish photographs from the flicker website and 50 artforms from the book, namely $Art\ forms\ in\ nature$ by Ernst Haeckel \cite{haeckel2012art}. The flicker website searched with the keyword $Jellyfish$. Search results are listed in order of relevance, and images with download permission are downloaded. Fifty artforms are collected. Sample drawings are given in Figure \ref{fig:artform_samples}.

The aerial-to-maps dataset contains 1096 satellite and maps image pairs scraped from google maps around New York City.

\vspace{-10pt}
\subsection{Qualitative Results}
\begin{figure}[]
\centering
    \scalebox{.2}{\includegraphics{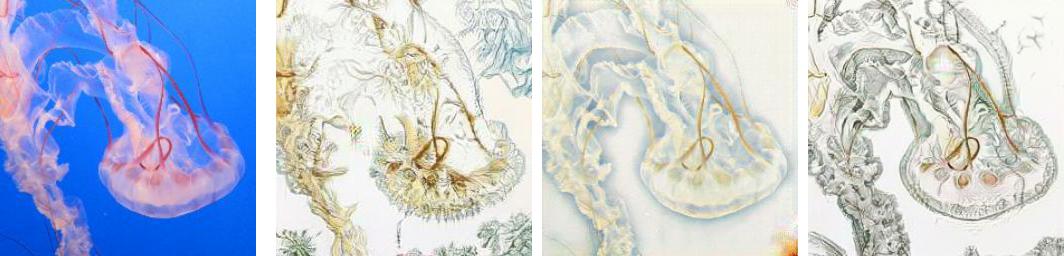}}
    \scalebox{.2}{\includegraphics{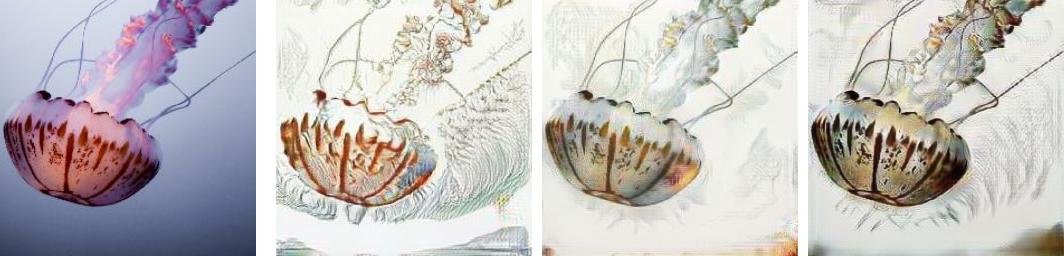}}
    \scalebox{.2}{\includegraphics{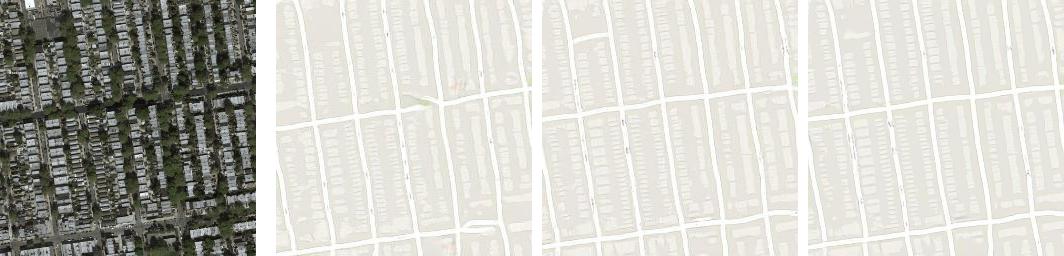}}
    \scalebox{.2}{\includegraphics{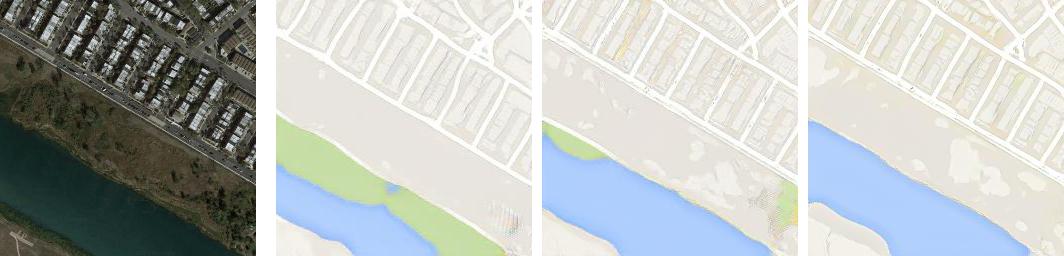}}
    \caption{The columns from left to right are original image, CycleGAN, GAIT ($c_{ga}=1$) and GAIT ($c_{ga}=2$) results for different images}
    \label{fig:qualitative_results}
\end{figure}

Images generated by the proposed method are shown in Figure \ref{fig:qualitative_results}. The columns in the figure show the original image, CycleGAN result, SobelGAN result on the Jellyfish-to-Haeckel dataset, respectively. As seen in the figure, the CycleGAN method managed to create sharp visuals, but it could not protect the uniform background information. The GAIT method proposed here can preserve the background with the help of the gradient adjustment loss function. Gradient adjustment constant $c_{ga}$ is taken as 1 in this experiments.

\vspace{-10pt}
\subsection{Quantitative Results}
The Kernel Inception Distance \cite{binkowski2018demystifying} is used to evaluate the proposed framework quantitatively. KID calculates the squared maximum mean discrepancy between the Inception network \cite{szegedy2016rethinking} features of real and generated images.

The gradient adjustment loss function helps conservation of the uniform background when $c_{ga}$ is taken as 1. However, KID scores decrease in this configuration, which corresponds to a decrease in marginal distribution matching performance. The sketch-like images have strong edges compared to the natural images. Therefore, it is safe to assume that gradient should be boosted when an image is translated from natural to sketch domain, such as jellyfish-to-Haeckel and Aerial-to-Maps. The KID scores on these datasets are shown in table \ref{tab:kid}. GAIT outperforms the CycleGAN method in the forward translation when $c_{ga}$ is selected as 2. The inverse translation is used as an auxiliary operation, and the performance increase is not aimed. Example images for two dataset with two different $c_{ga}$ values are shown in Figure \ref{fig:qualitative_results}. The inverse translation is an auxiliary operation to define the cyclic consistency loss function. 


\begin{table}[]
\centering
\caption{Kernel Inception Distance$\times$100 $\pm$ std.$\times$100. The lower value shows a higher similarity. J: Jellyfish, H:Haeckel, A: Aerial, M: Maps}
\label{tab:kid}
\begin{tabular}{lccc}
\hline
 & CycleGan & GAIT, $c_{ga}=1$ & GAIT, $c_{ga}=2$\\
\hline
J2H & 8.13$\pm$0.43 & 16.22$\pm$0.69 & \textbf{6.03}$\boldsymbol{\pm}$\textbf{0.38}\\
H2J & \textbf{3.48}$\boldsymbol{\pm}$\textbf{0.36} & 7.03$\pm$0.57& 4.62$\pm$0.3\\
A2M & 4.55$\pm$0.22 & 3.79$\pm$0.16 & \textbf{3.27}$\boldsymbol{\pm}$\textbf{0.15}\\
M2A & \textbf{1.62}$\boldsymbol{\pm}$\textbf{0.14} & 13.40$\pm$0.48 & 19.37$\pm$0.57\\
\hline
\end{tabular}
\end{table}

\vspace{-10pt}

\section{Conclusion}
Current unsupervised IIT methods can create artifacts in the uniform background of the source image where the source and target domains have different background distributions. The gradient of the original image is preserved during the translation, and the defects in the uniform background are eliminated with the help of the gradient adjustment loss function. In addition, the distribution matching capability is enhanced by the help of the gradient adjustment constant when the gradient is boosted. The results are presented on the Jellyfish-to-Haeckel and Aerial-to-Maps datasets, and the effectiveness of the proposed method is demonstrated with qualitative and quantitative results. The results are compared with the CycleGAN method, and a performance gain is observed.

\bibliographystyle{IEEEbib}
\bibliography{refs}

\begin{thebibliography}{10}

\bibitem{yi2017dualgan}
Zili Yi, Hao Zhang, Ping Tan, and Minglun Gong,
\newblock ``Dualgan: Unsupervised dual learning for image-to-image
  translation,''
\newblock in {\em Proceedings of the IEEE international conference on computer
  vision}, 2017, pp. 2849--2857.

\bibitem{kim2017learning}
Taeksoo Kim, Moonsu Cha, Hyunsoo Kim, Jung~Kwon Lee, and Jiwon Kim,
\newblock ``Learning to discover cross-domain relations with generative
  adversarial networks,''
\newblock in {\em Proceedings of the 34th International Conference on Machine
  Learning-Volume 70}. JMLR. org, 2017, pp. 1857--1865.

\bibitem{zhu2017unpaired}
Jun-Yan Zhu, Taesung Park, Phillip Isola, and Alexei~A Efros,
\newblock ``Unpaired image-to-image translation using cycle-consistent
  adversarial networks,''
\newblock in {\em Proceedings of the IEEE international conference on computer
  vision}, 2017, pp. 2223--2232.

\bibitem{goodfellow2014generative}
Ian Goodfellow, Jean Pouget-Abadie, Mehdi Mirza, Bing Xu, David Warde-Farley,
  Sherjil Ozair, Aaron Courville, and Yoshua Bengio,
\newblock ``Generative adversarial nets,''
\newblock in {\em Advances in neural information processing systems}, 2014, pp.
  2672--2680.

\bibitem{radford2015unsupervised}
Alec Radford, Luke Metz, and Soumith Chintala,
\newblock ``Unsupervised representation learning with deep convolutional
  generative adversarial networks,''
\newblock {\em arXiv preprint arXiv:1511.06434}, 2015.

\bibitem{mao2017least}
Xudong Mao, Qing Li, Haoran Xie, Raymond~YK Lau, Zhen Wang, and Stephen
  Paul~Smolley,
\newblock ``Least squares generative adversarial networks,''
\newblock in {\em Proceedings of the IEEE International Conference on Computer
  Vision}, 2017, pp. 2794--2802.

\bibitem{isola2017image}
Phillip Isola, Jun-Yan Zhu, Tinghui Zhou, and Alexei~A Efros,
\newblock ``Image-to-image translation with conditional adversarial networks,''
\newblock in {\em Proceedings of the IEEE conference on computer vision and
  pattern recognition}, 2017, pp. 1125--1134.

\bibitem{mejjati2018unsupervised}
Youssef~Alami Mejjati, Christian Richardt, James Tompkin, Darren Cosker, and
  Kwang~In Kim,
\newblock ``Unsupervised attention-guided image-to-image translation,''
\newblock in {\em Advances in Neural Information Processing Systems}, 2018, pp.
  3693--3703.

\bibitem{liu2016coupled}
Ming-Yu Liu and Oncel Tuzel,
\newblock ``Coupled generative adversarial networks,''
\newblock in {\em Advances in neural information processing systems}, 2016, pp.
  469--477.

\bibitem{liu2017unsupervised}
Ming-Yu Liu, Thomas Breuel, and Jan Kautz,
\newblock ``Unsupervised image-to-image translation networks,''
\newblock in {\em Advances in neural information processing systems}, 2017, pp.
  700--708.

\bibitem{huang2018multimodal}
Xun Huang, Ming-Yu Liu, Serge Belongie, and Jan Kautz,
\newblock ``Multimodal unsupervised image-to-image translation,''
\newblock in {\em Proceedings of the European Conference on Computer Vision
  (ECCV)}, 2018, pp. 172--189.

\bibitem{johnson2016perceptual}
Justin Johnson, Alexandre Alahi, and Li~Fei-Fei,
\newblock ``Perceptual losses for real-time style transfer and
  super-resolution,''
\newblock in {\em European conference on computer vision}. Springer, 2016, pp.
  694--711.

\bibitem{haeckel2012art}
Ernst Haeckel,
\newblock {\em Art forms in nature},
\newblock Courier Corporation, 2012.

\bibitem{binkowski2018demystifying}
Miko{\l}aj Bi{\'n}kowski, Dougal~J Sutherland, Michael Arbel, and Arthur
  Gretton,
\newblock ``Demystifying mmd gans,''
\newblock {\em arXiv preprint arXiv:1801.01401}, 2018.

\bibitem{szegedy2016rethinking}
Christian Szegedy, Vincent Vanhoucke, Sergey Ioffe, Jon Shlens, and Zbigniew
  Wojna,
\newblock ``Rethinking the inception architecture for computer vision,''
\newblock in {\em Proceedings of the IEEE conference on computer vision and
  pattern recognition}, 2016, pp. 2818--2826.

\end{thebibliography}

\end{document}